\documentclass{article}
\usepackage{amsmath,spconf}

\usepackage{times}
\usepackage{epsfig}
\usepackage{graphicx}
\usepackage{subcaption}
\usepackage{amsmath}
\usepackage{amssymb}
\usepackage{textcomp}
\usepackage{soul}
\usepackage{changes}
\usepackage{multirow}
\usepackage{xcolor}

\usepackage[breaklinks=true,bookmarks=false]{hyperref}

\usepackage{ulem}

\begin{document}

\toappear{2019 IEEE International Workshop on Machine Learning for Signal Processing, Oct.\ 13--16, 2019, Pittsburgh, PA, USA}

\copyrightnotice{978-1-7281-0824-7/19/\$31.00 {\copyright}2019 IEEE}
	
\title{VAE/WGAN-Based Image Representation Learning for Pose-Preserving Seamless Identity Replacement in Facial Images}

\name{Hiroki Kawai, Jiawei Chen, Prakash Ishwar, Janusz Konrad}
\address{Department of Electrical and Computer Engineering, Boston Univeristy, Boston, USA \\ Email: \{hirokik, garychen, pi, jkonrad\}@bu.edu}

\maketitle
\setstcolor{red}
\begin{abstract}
	We present a novel variational generative adversarial network (VGAN) based on Wasserstein loss to learn a latent representation from a face image that is invariant to identity but preserves head-pose information. This facilitates synthesis of a realistic face image with the same head pose as a given input image, but with a different identity. One application of this network is in privacy-sensitive scenarios; after identity replacement in an image, utility, such as head pose, can still be recovered. Extensive experimental validation on synthetic and real human-face image datasets performed under 3 threat scenarios confirms the ability of the proposed network to preserve head pose of the input image, mask the input identity, and synthesize a good-quality realistic face image of a desired identity. We also show that our network can be used to perform pose-preserving identity morphing and identity-preserving pose morphing. The proposed method improves over a recent state-of-the-art method in terms of quantitative metrics as well as synthesized image quality.
\end{abstract}

\section{Introduction}

Sensor-equipped, algorithm-driven smart living spaces of the future promise to deliver increased energy efficiency, health benefits, and productivity gains~\cite{chen2016estimating,chen2017semi,roeper2016privacy,zhao2017privacy}. 
This will require recognition of occupant's activities, gestures, body pose, facial expressions, etc. While it can be accomplished using video cameras, their use will likely raise privacy concerns which can hinder further development~\cite{Erdelyi2018}. 
One approach to address privacy concerns is to obscure visually-identifying information {\it via} explicit image manipulation, such as pixelization, blurring, cartooning, etc. However, such manipulation may also obscure utility information, such as facial expression or gesture. An alternative approach is to seamlessly alter occupant's identity in an image without significantly degrading image quality and utility information. This idea was first proposed for facial expression classification in Privacy-Preserving Representation-Learning VGAN (PPRL-VGAN) framework \cite{Chen_2018_CVPR_Workshops}. Specifically, a VGAN was trained to learn an identity-invariant latent representation of an input face image from which a realistic image could be synthesized with the same facial expression as the input image, but a different, user-specified identity. 
\begin{figure}[!htb]
	\vglue -0.3cm
	\centering{\includegraphics[width=1.0\linewidth]{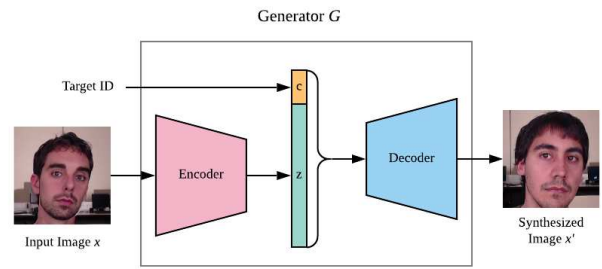}}
	\vglue -3ex
	\caption{\small Pose-preserving identity replacement in a human-head image. An input image is encoded into an identity-invariant latent representation $z$, which is then concatenated with target identity information $c$ and fed into the decoder to generate an identity-altered image while preserving the original head pose.}
	\label{fig:top}
\end{figure}

In this paper, we extend the PPRL-VGAN framework to head-pose estimation and introduce several innovations in its architecture and training. Specifically, this paper makes the following contributions:
\begin{enumerate}
	\item 
	We extend PPRL-VGAN that was developed for facial expression classification (a discrete attribute), to head-pose estimation (a continuous attribute). This requires learning regression functions in addition to classification functions.
	
	\item 
	We substantially modify the PPRL-VGAN architecture to stabilize the training process and improve pose estimation accuracy (customized na\"ive inception modules \cite{43022}, gradient-penalty Wasserstein GAN cost function \cite{pmlr-v70-arjovsky17a}, Adam optimizer \cite{journals/corr/KingmaB14}).
	
	\item We introduce an additional image-reconstruction loss term in the generator's cost function to improve the visual quality of generated images.
	
	\item We provide quantitative and qualitative validation of our approach on synthetic and real-world datasets under three different privacy-threat scenarios. 
	
	\item We demonstrate our method's capability to perform pose-preserving identity morphing and identity-preserving pose morphing of facial images by interpolating, respectively, the identity code and latent representation within the generator.   
\end{enumerate}

\section{Related Work}
\noindent \textbf{Privacy-Preserving Head Pose Estimation:}
There is an extensive body of literature devoted to image-based human head pose estimation spanning several decades.
Focusing on identity-invariant head pose estimation, early works such as \cite{Sherrah99, Sherrah2001} used classical statistical learning approaches that emphasized differences between head poses while suppressing differences between identities.
%
In more recent work, \cite{martin2014toward} develops a deidentification filter on video sequences of car drivers and then estimates a driver's gaze, head dynamic, etc. In~\cite{chen2016estimating}, a non-linear regression method is proposed to estimate human head pose from extremely low resolution images in which identity is visually imperceptible.
However, the aforementioned methods either only produce coarse-grained head pose estimates~\cite{Sherrah99,Sherrah2001,martin2014toward} or perform below par in head pose estimation~\cite{chen2016estimating}.

\noindent \textbf{Invariant Representation Learning:}
Invariant representation learning has been extensively studied in various contexts. 
%
%
Recent studies~\cite{edwards2015censoring,lample2017fader,xie2017controllable} utilize VAEs with adversarial training to learn an invariant latent space which also enables generating new data samples.
Our method also leverages VAEs and GANs, but unlike these works we apply adversarial training in the image space instead of the latent space. This creates better quality images.
Particularly relevant to our work is the PPRL-VGAN proposed in~\cite{Chen_2018_CVPR_Workshops} to learn an identity-invariant face image representation that preserves a discrete set of facial expressions. 
%
%
Our proposed framework is similar to PPRL-VGAN in the use of a VGAN, but significantly differs from it in ways described above in the list of contributions.
%
%
%
%
%
%
%

\section{Background Material}

%
\noindent \textbf{Variational Auto-Encoder (VAE):} consists of an \textit{encoder} (Enc) network and a \textit{decoder} (Dec) network. The encoder maps a given data sample $\mathbf{x}$ into a lower-dimensional latent representation $\mathbf{z}$. The decoder maps $\mathbf{z}$ back to data space:
\begin{equation*}
\mathbf{z} \sim Enc(\mathbf{x})=q(\mathbf{z}|\mathbf{x}), \quad
%
\hat{\mathbf{x}} \sim Dec(\mathbf{z})=p(\mathbf{x}|\mathbf{z})
\end{equation*}
The encoder and decoder are jointly trained to minimize a VAE loss function which is a combination of a reconstruction term and a prior regularization term:
\begin{equation*}
\!\!\!\!\! L^{VAE}_\mathbf{x} = -E_{q(\mathbf{z}|\mathbf{x})}[\text{log}(p(\mathbf{x}|\mathbf{z}))] + D_{KL}(q(\mathbf{z}|\mathbf{x})||p(\mathbf{z}))
\end{equation*}
where $\mathbf{z}\sim p(\mathbf{z}) = \mathcal{N}(\mathbf{0}, \mathbf{I})$ is a prior for the latent distribution which regularizes the encoder training and $D_{KL}$ is the Kullback-Leibler divergence. 

\noindent \textbf{Generative Adversarial Network:}
%
A plain GAN consists of a generator network ($G$) and a discriminator  network ($D$) that are trained by a min-max game competition. 
Whereas $G$ adjusts its weights to map a source of noise $\mathbf{z} \sim p_z(\mathbf{z})$ to the data space, $D$ aims to reliably distinguish real data samples $\mathbf{x}\sim p_d(\mathbf{x})$ from fake data samples $G(\mathbf{z})$.
The output from the discriminator can be interpreted as a probability that the input data is real.
In practice, the two networks are optimized in an alternating manner {\it via} a loss function:
\begin{equation*}
\begin{aligned}
\!\!\!\!\!\!\min_G \max_D \underbrace{\tiny E_{\mathbf{x}\sim p_d(\mathbf{x})}[\log{D(\mathbf{x})}]
	+E_{\mathbf{z}\sim p_z(\mathbf{z})}[\log(1-D(G(\mathbf{z})))]}_{L(G,D)}    
\end{aligned}
\end{equation*}
\noindent \textbf{Wasserstein GAN (WGAN) \cite{pmlr-v70-arjovsky17a}:} \label{WGAN}
WGAN is an important extension of GAN which improves image quality and stabilizes training. 
In the plain GAN formulation, optimizing the networks according to the minimax objective is equivalent to minimizing the Jensen-Shannon (JS) divergence between the data distribution $p_d$ and the distribution parameterized by the generator network $p_g$.
The JS divergence is, however, potentially discontinuous in the generator's parameters which makes training difficult~\cite{NIPS2017_7159}.
%
%
WGAN instead proposes to minimize the Wasserstein distance $W(p_d, p_g)$, which is a smooth function of network parameters (under minor technical conditions):
\begin{equation*}
\begin{aligned}
W(p_d, p_g) = \!\!\!\!\!\! \max_{\omega : D_\omega(\mathbf{x}) \in F_1} \!\! E_{\mathbf{x}\sim p_d}&[D_\omega(\mathbf{x})]-
E_{\mathbf{x}\sim p_g}[D_\omega(G(\mathbf{z}))]
\end{aligned}
\end{equation*}
where $F_1$ is the set of 1-Lipshitz functions. 
In order to enforce 1-Lipshitz continuity, we follow  previous work~\cite{NIPS2017_7159} and a gradient penalty term  $\gamma E_{\mathbf{x}\sim p_h(\mathbf{x})}[(||\nabla_\mathbf{x} D_\omega(\mathbf{x})||_2-1)^2]$  to the cost function, where $\gamma > 0$ is a tuning parameter, $p_h(\mathbf{x}) = c p_d(\mathbf{x})+(1-c)p_g(\mathbf{x})$ and $c \in [0, 1]$.
%
%

\begin{figure*}[!t]
	\centering{\includegraphics[width=0.95\linewidth]{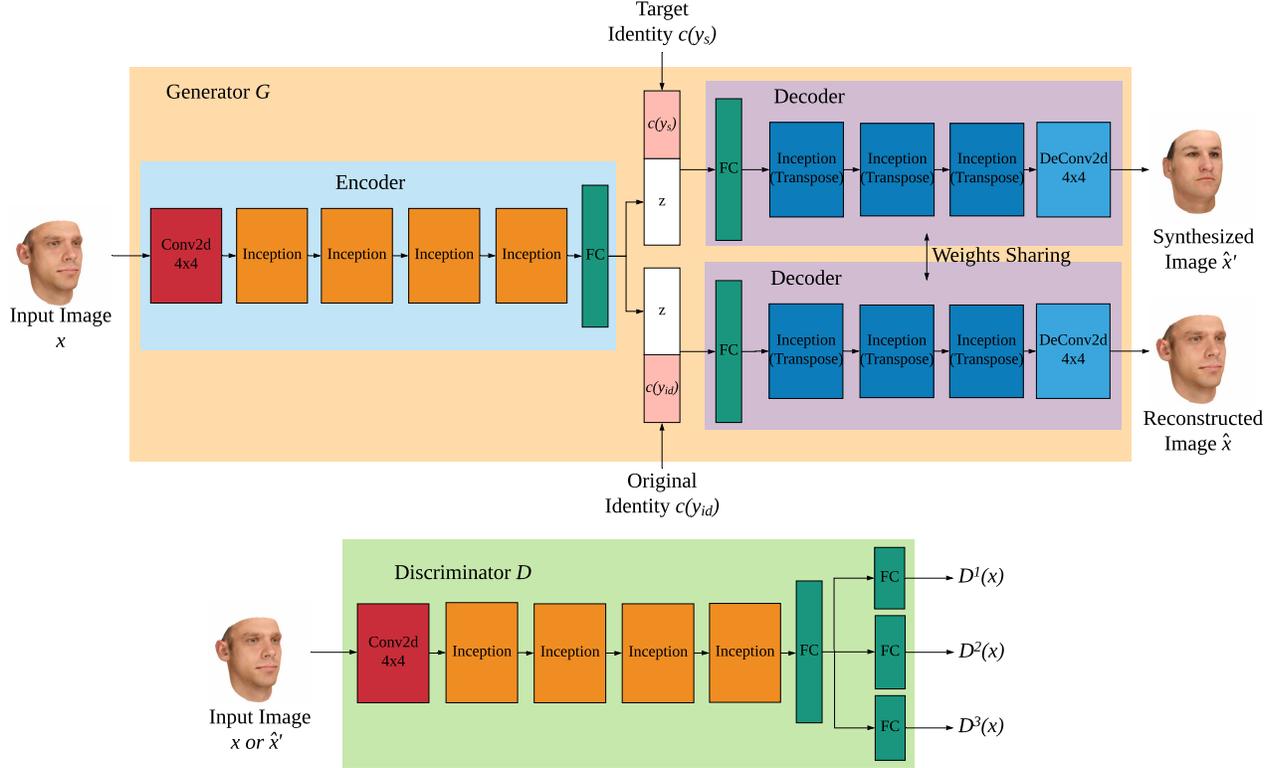}}
	\vglue -8ex
	\caption{\small Proposed VAE WGAN. Training alternates between optimizing $G$ and $D$. Each inception module performs convolution on an input tensor with filters of sizes 1$\times$1, 2$\times$2, and 4$\times$4, and the results are concatenated to produce an output tensor.}
	\vglue -0.1in
\end{figure*}
\section{Proposed Methodology}
%
%
%
%
%

Let $\mathbf{x}$ be a given face image with one (discrete) identity label $y_{id} \in \{1,..., N_s\}$, where $N_s$ is the number of distinct subjects, and three (fine-grained continuous)
head pose labels $\mathbf{y}_{pose} = \{y^1_{pose}, y^2_{pose}, y^3_{pose}\}$ corresponding to the yaw, pitch and roll angles respectively. Our proposed framework has two objectives:
1) to produce a low-dimensional image representation $\mathbf{z}$ which is invariant to identity information but retains the head pose attributes, and 
2) to synthesize a realistic face image with the same head pose as the input image, but a target identity specified by an input identity code $\mathbf{c}(y_s)$, where $y_s \in \{1,..., N_s\}$ is generated from a distribution $p(y_s)$ and $\mathbf{c}(\cdot)$ is the one-hot encoding function.

The generator in our model is structured similarly to a VAE. 
%
The encoder network encodes an input face image $\mathbf{x}$ to a representation $\mathbf{z} \sim Enc(\mathbf{x})$ and the decoder 
maps the latent representation $\mathbf{z}$ in combination with an identity code $\mathbf{c}$ back to the image space. 
The discriminator $D$ consists of three networks: $D_{\omega}^1, D^2$ and $D^3$. 
The $D_{\omega}^1$ network is a detector which is trained to predict if the input image is real or generated,
%
%
the $D^2$ network is a classifier which is trained to recognize the identity of the person in the input image, and 
the $D^3$ network is a regressor which is trained to estimate the three head pose angles of the input image.

%
For identity $y_{id}$, the decoder is trained to generate an accurate reconstruction of the input $\hat{\mathbf{x}} \sim Dec(\mathbf{z}, \mathbf{c}(y_{id}))$.
For a randomly sampled identity $y_s$, the decoder is also trained to synthesize a realistic face image $\hat{\mathbf{x}}' \sim Dec(\mathbf{z}, \mathbf{c}(y_{s}))$ so that discriminator $D^2$ classifies it as the target identity $y_s$ while simultaneously ensuring that regressor $D^3$ correctly estimates the head pose of the input image $\mathbf{x}$.
%
%
Specifically, the generator's weights are updated to \textit{minimize} the cost
%
\begin{equation}\label{loss_g}
\begin{aligned}
L_G &= E_{(\mathbf{x}, y_{id}, \mathbf{y}_{pose}) \sim p_d(\mathbf{x}, y_{id}, \mathbf{y}_{pose}), y_s \sim p(y_s)} \big[ \\
& - \lambda_1^G D_{\omega}^{1}(G(\mathbf{x}, \mathbf{c}(y_s)))-\lambda_2^G\log(D_{y_s}^2(G(\mathbf{x}, \mathbf{c}(y_s)))) +  \\
& \lambda_3^G\sum_{i= 1}^{3}|y^{i}_{pose}-D_{i}^3\big(G(\mathbf{x}, \mathbf{c}(y_s))\big)| + \\
&\lambda_4^G||G(\mathbf{x}, \mathbf{c}(y_{id}))-\mathbf{x}||_2^2
+ \lambda_5^G D_{KL}(q(\mathbf{z}|\mathbf{x})||r(\mathbf{z})) \big]
\end{aligned}
\end{equation}
%
where $D_{\omega}(\mathbf{x})\in F_1$,
$D_i^2$ is the predicted probability of the $i$-th identity, $D_i^3$ ($i \in \{1,2,3\}$) are the predicted pose angles, $\lambda_i^G$ ($i\in \{1,2,3,4,5\}$) are tuning factors, and $r(\mathbf{z}) \sim \mathcal{N}(\mathbf{0}, \mathbf{I})$ is the prior distribution of the latent representation.
%
In comparison to the PPRL-VGAN generator cost function, ours adopts the WGAN formulation for $D^1$ to stabilize training. Furthermore, $D^3$ is optimized to accurately estimate the (continuous) head pose angles from an input image as opposed to recognizing facial expressions.
For the pose-angle loss, we used the $L1$ norm of the difference between the predicted pose angles and the ground-truth pose angles. There is no need for angle unwrapping since the datasets used in our experiments contain face images with head pose angles limited to $[-90^{\circ}, 90^{\circ}]$. 
We acknowledge that alternative loss functions could be used for penalizing pose estimation error, e.g., geodesic loss, however comparing different loss functions is out of the scope of this work. 
Lastly, the additional $L2$ reconstruction error term (the term attached to $\lambda_4^G$)  assists the model in generating good-quality images.

The discriminator is optimized to maximize the dual form of the Wasserstein distance between the real data distribution and the generator's distribution. 
When given a real training sample, it is also trained to accurately recognize the person's identity and head pose.
This is accomplished by updating $D$'s weights to \textit{maximize} the cost 
%
\begin{equation}\label{loss_d}
\begin{aligned}
L_D &= \lambda_1^D \{E_{\mathbf{x} \sim p_d(\mathbf{x})} [D_{\omega}^1(\mathbf{x})] - \\
& E_{\mathbf{x} \sim p_d(\mathbf{x}), y_s \sim p(y_s)} [D_{\omega}^1(G(\mathbf{x}, \mathbf{c}(y_s)))]\} + \\
& E_{(\mathbf{x}, y_{id}, \mathbf{y}_{pose}) \sim p_d(\mathbf{x}, y_{id}, \mathbf{y}_{pose})}[\lambda_2^D\log(D_{y_{id}}^2(\mathbf{x})) - \\
& \lambda_3^G\sum_{i=1}^{3}|y^{i}_{pose}-D_{i}^3(\mathbf{x})| \ ] - \\
& \lambda_4^D E_{\mathbf{x}\sim p_h(\mathbf{x})}[(||\nabla_\mathbf{x} D_{\omega}^{1}(\mathbf{x})||_2-1)^2]
\end{aligned}
\end{equation}
where $\lambda_i^D$ ($i \in \{1,2,3,4\}$) are tunable weighting factors. 
The last term in Eq.~(\ref{loss_d}) punishes the gradient to ensure that $D_{\omega}^{1}$ is a 1-Lipschitz function of $\mathbf{x}$.
A significant point of difference compared to PPRL-VGAN is that our discriminator's cost function leverages the WGAN formulation via a gradient-penalty on $D_{\omega}^1$. Another major difference is our use of a (continuous) head pose estimation loss in lieu of the (discrete) expression recognition loss used in PPRL-VGAN.

The weights of $G$ and $D$ are updated in alternating order. Over successive rounds of training, the generator learns to fit the real data distribution and synthesize images that can fool $D$.
As the input code $\mathbf{c}(y_s)$ determines the identity of the synthesized image, the encoder is encouraged to eliminate information about the identity of $\mathbf{x}$ in the latent space.
Moreover, as $\hat{\mathbf{x}}'$ must preserve the head pose of $\mathbf{x}$, the encoder is also encouraged to embed head pose attributes within the latent space.
The reconstruction objective in Eq.~(\ref{loss_g}) additionally encourages the encoder to pass head pose information to the latent space and promotes synthesis of good quality images. 

\vspace{-1ex}
\section{Experimental Evaluation}\label{evaluation}
We evaluated the performance of the proposed model on the UPNA head pose dataset and its synthetic replica~\cite{ariz2016novel,larumbe2017improved}.
%
Both datasets contain 10 videos for each of 10 subjects. In total, one dataset includes 35,990 frames. Ground-truth continuous head pose angles (yaw, pitch, roll) and a face-centered bounding box are provided for each frame. 
In our experiments, we first cropped each frame using the provided bounding box and then resized it to 64$\times$64-pixel resolution. 
For each subject, we selected 80$\%$ of the frames from each video for training and used the remaining 20\% for testing.

We compared our model with PPRL-VGAN~\cite{Chen_2018_CVPR_Workshops} which was modified to preserve continuous head poses 
%
by replacing the facial expression classifier in the discriminator with a head pose estimator and changing both generator and discriminator cost functions to encourage preservation of head pose information within the latent and synthetic image spaces.  

\subsection{Quantitative evaluation}
We used the methodology in~\cite{Chen_2018_CVPR_Workshops} to evaluate pose-preserving identity replacement performance under three threat scenarios.

\noindent \textbf{Attack scenario I:}
This is the least privacy threatening case wherein the attacker has access to the original training images with their ground-truth labels $(\mathbf{x}_{train}, y_{id}^{train})$. 
However, all the test images have been passed through the trained model with a randomized identity $y_s$ for privacy protection.
The attacker can train an identifier using the unaltered training set and apply it to the privacy-protected test images $\hat{\mathbf{x}}'_{test} = G(\mathbf{x}_{test}, \mathbf{c}(y_s))$ to predict their underlying ground-truth identities $y_{id}^{test}$.
%

\noindent \textbf{Attack scenario II:}
This is a more challenging scenario 
in which the attacker can access the privacy-protected training images $\hat{\mathbf{x}}'_{train} = G(\mathbf{x}_{train}, \mathbf{c}(y_s))$ that have been processed by the trained model, and the attacker knows the corresponding ground-truth identity labels. 
As a result, the attacker can train an identifier on the training set that has been protected using the same transformation as the test set.  
It is possible that attacker's identifier can uncover the underlying identity of a protected test image if the proposed model
fails to eliminate information about the original identity in the synthesized image.
%

\noindent \textbf{Attack scenario III:}
Here the attacker has access to our model and thus can obtain the latent representation $\mathbf{z}$ for a given image $\mathbf{x}$. 
Therefore, the attacker can train an identifier using $(\mathbf{z}_{train}, y_{id}^{train})$ and apply it to the representation $\mathbf{z}_{test}$ of a test image to predict its identity. 

%

In order to assess how well pose is preserved in the synthesized images, we trained a dedicated head pose estimator for each scenario using the available type of training images and their ground-truth head pose labels.  
We then applied the trained estimator to the test images to measure the head pose estimation performance. 

We use correct classification rate (CCR) and mean absolute error (MAE) to measure the performance of identification and head pose estimation, respectively.
In all three scenarios, a low identification CCR and a small head pose estimation error are favored.
%
%
%
\begin{table}[!ht]
	\begin{center}
		\resizebox{0.9\linewidth}{!}{
			\begin{tabular}{|c|c|c|c|c|}
				\hline
				\multirow{3}{*}{Scenarios}           & \multicolumn{2}{c|}{Identification(\%)}& \multicolumn{2}{c|}{MAE Average(\textdegree)} \\ \cline{2-5}
				& \multirow{2}{*}{Ours} &PPRL- & \multirow{2}{*}{Ours} &PPRL-  \\
				& &VGAN & &VGAN \\
				\hline
				Privacy Unconstrained & \multicolumn{2}{c|}{99.97} & \multicolumn{2}{c|}{0.69} \\ \hline
				Attack Scenario I     &  10.23 & {9.92} & {2.251} & 3.57 \\ \hline
				Attack Scenario II    & 23.31 & {21.64} & {1.811} & 2.90 \\ \hline
				Attack Scenario III   & {21.33} & 23.71 & {2.212} & 2.76 \\ \hline
			\end{tabular}
		}
	\end{center}
	\vglue -2ex
	\caption{\small Classification CCRs for person identification and MAE for head pose estimation on UPNA. 
	}
	\vglue -2ex
	\label{Tab:synthetic_uvsp}
\end{table}
\begin{table}[!ht]
	\begin{center}
		\resizebox{0.9\linewidth}{!}{
			\begin{tabular}{|c|c|c|c|c|}
				\hline
				\multirow{3}{*}{Scenarios}           & \multicolumn{2}{c|}{Identification(\%)}& \multicolumn{2}{c|}{MAE Average(\textdegree)} \\ \cline{2-5}
				& \multirow{2}{*}{Ours} &PPRL- & \multirow{2}{*}{Ours} &PPRL-  \\
				& &VGAN & &VGAN \\
				\hline
				Privacy Unconstrained & \multicolumn{2}{c|}{100.00} & \multicolumn{2}{c|}{0.60} \\ \hline
				Attack Scenario I     & {10.06} & 10.47 & {2.27} & 5.68 \\ \hline
				Attack Scenario II    & 26.51 & {17.36} & {1.74} & 3.65 \\ \hline
				Attack Scenario III   & {24.49} & 25.16 & {2.10} & 2.77 \\ \hline
			\end{tabular}
		}
	\end{center}
	\vglue -2ex
	\caption{\small Classification CCRs for person identification and MAE for head pose estimation on UPNA synthetic.
	}
	\vglue -4ex
	\label{Tab:realistic_uvsp}
\end{table}
The identification and head pose estimation performance of the two competing models under various scenarios are reported in  Table~\ref{Tab:synthetic_uvsp} for the UPNA dataset and Table~\ref{Tab:realistic_uvsp} for its synthetic version.
In the privacy unconstrained scenario, both training and testing data are unaltered. 
The resulting identification CCRs lower-bound the attainable identification accuracy while the resulting head pose estimation MAEs upper-bound the attainable estimation error.
%
%
%
In attack scenario I, we observe that the identification performance of both models is close to pure chance (10$\%$ for both datasets since each includes 10 subjects). This indicates both methods succeed in protecting identity when the attacker has no knowledge about the applied privacy-protection transformation.  
%
As for head-pose estimation, we can see our model consistently outperforms the benchmark method by 1-3 degrees. 
In attack scenario II, the identification CCRs of both methods are close but higher than those in attack scenario I by 7-17$\%$.
This suggests a certain amount of the identity information has leaked into the synthesized images.
However, the resulting CCRs are still much lower than those in the privacy unconstrained scenario.
In terms of preserving head pose, the proposed model again outperforms the benchmark method on the two datasets, more than halving the error. 
As for attack scenario III, the identification and pose estimation performance of both methods are similar.
%
Overall, the proposed model outperforms the benchmark method in preserving the head pose information and has comparable performance in terms of disentangling identity information.
%
%

\begin{figure*}
	\centering{\includegraphics[width=0.95\linewidth]{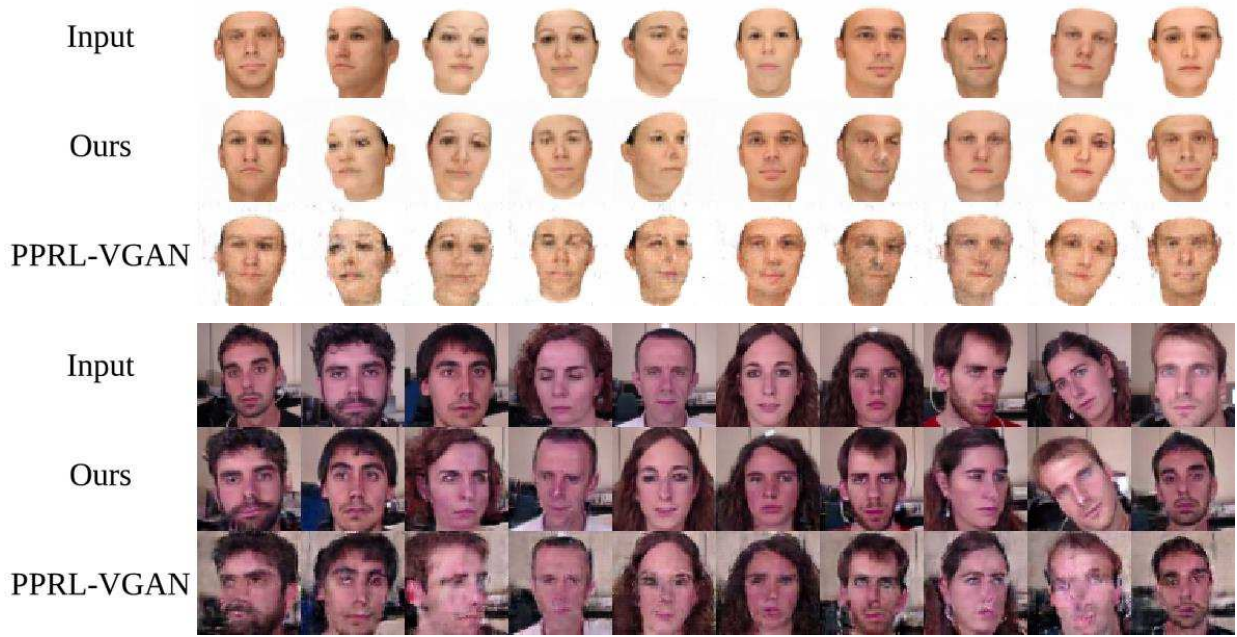}}
	\vglue -6ex 
	\caption{\small Pose-preserving identity replacement on the UPNA synthetic (top) and UPNA (bottom) datasets. In each sub-figure, the first row shows one example input image for each identity. The second and third rows show images synthesized (for the same target identity) from the proposed model and the benchmark model, respectively. 
	}
	\vglue -2ex 
	\label{fig:synthesized}
\end{figure*}
\subsection{Qualitative evaluation}
\noindent \textbf{Identity replacement:}
Once trained, our model can synthesize a new face image (see Fig.~\ref{fig:synthesized}) with the same head pose as the input image and a target identity specified by the identity code $\mathbf{c}(y_s)$. 
Compared to the images generated by PPRL-VGAN, the synthetic images from our model
have better visual quality (e.g., contain fewer artifacts) and preserve the head pose attribute more accurately. 
%

\begin{figure*}[t!]
	\begin{center}
		\includegraphics[width=0.9\linewidth]{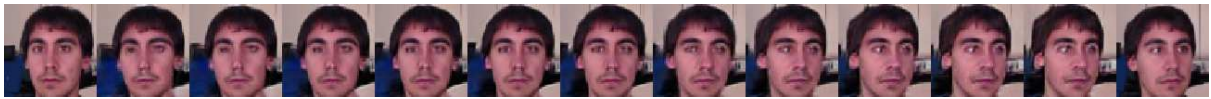}
	\end{center}
	\vglue -3ex
	\caption{\small Identity-preserving  pose morphing. The left-most and right-most images contain the initial and final poses of the same identity. The intermediate images are synthesized by linearly interpolating their latent representations.
	}
	\vglue -2ex
	\label{fig:interpolation}
\end{figure*}
\begin{figure*}[t!]
	\begin{center}
		\includegraphics[width=0.9\linewidth]{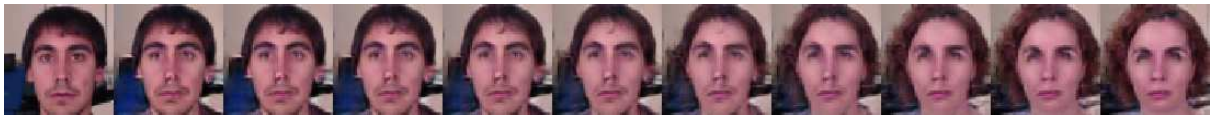}
	\end{center}
	\vglue -3ex
	\caption{\small Pose-preserving identity morphing. 
		The left-most and right-most images contain the initial and final identities in the same pose. The intermediate images are synthesized by linearly interpolating their identity codes.
	}
	\vglue -4ex
	\label{fig:morphing}
\end{figure*}
\noindent \textbf{Interpolating latent representations (pose morphing):}
In order to further evaluate the generative capability of the proposed model, we conducted additional experiments for identity-preserving pose morphing:
given a pair of source images $\mathbf{x}_{\mathrm{initial}}$, $\mathbf{x}_{\mathrm{final}}$ with the same identity but different head pose, and their corresponding latent representations $\mathbf{z}_{\mathrm{initial}}$, $\mathbf{z}_{\mathrm{final}}$, we linearly interpolate between $\mathbf{z}_{\mathrm{initial}}$ and $\mathbf{z}_{\mathrm{final}}$ to generate new representations as follows:
%
\begin{align*}
\mathbf{z}_{interp} = k\mathbf{z}_{\mathrm{initial}} + (1-k)\mathbf{z}_{\mathrm{final}}, \quad k\in [0,1].
\label{eq:rep_interpolation}
\end{align*}
Then, we synthesize new images by passing $(\mathbf{c}(y_{id}), \mathbf{z}_{interp})$ to the decoder. We observe that the head pose of synthesized images changes smoothly with $k$ (see Fig.~\ref{fig:interpolation}), suggesting that our model can capture salient head pose characteristics in the latent space.

\noindent \textbf{Interpolating identity codes (identity morphing):}
We also linearly interpolated between two identity codes $\mathbf{c}_{\mathrm{intial}}$ and $\mathbf{c}_{\mathrm{final}}$ to create new identity codes $\mathbf{c}_{\mathrm{interp}}$ as follows:
\begin{align*}
\mathbf{c}_{\mathrm{interp}} = k\mathbf{c}_{\mathrm{initial}} + (1-k)\mathbf{c}_{\mathrm{final}}, \quad k\in [0,1].
\end{align*}
Then, we passed the generated code with a fixed image representation (capturing pose) to the decoder. Interestingly, our trained model can generate a sequence of face images that exhibit a seamless transition from the initial identity to a target identity, i.e., face morphing (see Fig.~\ref{fig:morphing}), despite the fact that the model can only see one-hot codes specifying a \textit{discrete} set of identities during training.

\section{Conclusion}
We presented a framework for learning an identity-invariant image representation which retains fine-grained head pose attributes.
Quantitative results show that our model outperforms a recent state-of-the-art method for learning an identity-invariant image representation. 
Our model also enables synthesis of a realistic face image with a desired identity. 
Finally, our model can be applied to other image tasks such as pose or face morphing. 


{\small
	\bibliographystyle{ieee}
	\bibliography{bibliography}
}
\end{document}